# DSBI: Double-Sided Braille Image Dataset and Algorithm Evaluation for Braille Dots Detection

Renqiang Li
Institute of Computing Technology, Chinese Academy of Sciences, Beijing, China
lirenqiang@ict.ac.cn

Hong Liu*
Institute of Computing Technology, Chinese Academy of Sciences, Beijing, China
hliu@ict.ac.cn

Xiangdong Wang
Institute of Computing Technology, Chinese Academy of Sciences, Beijing, China
xdwang@ict.ac.cn

Yueliang Qian
Institute of Computing Technology, Chinese Academy of Sciences, Beijing, China
ylqian@ict.ac.cn

## ABSTRACT

Braille is an effective way for the visually impaired to learn knowledge and obtain information. Braille image recognition aims to automatically detect Braille dots in the whole Braille image. There is no available public datasets for Braille image recognition to push relevant research and evaluate algorithms. This paper constructs a large-scale Double-Sided Braille Image dataset DSBI with detailed Braille recto dots, verso dots and Braille cells annotation. To quickly annotate Braille images, an auxiliary annotation strategy is proposed, which adopts initial automatic detection of Braille dots and modifies annotation results by convenient human-computer interaction method. This labeling strategy can averagely increase label efficiency by six times for recto dots annotation in one Braille image. Braille dots detection is the core and basic step for Braille image recognition. This paper also evaluates some Braille dots detection methods on our dataset DSBI and gives the benchmark performance of recto dots detection. We have released our Braille images dataset on the GitHub website.

## 1. INTRODUCTION

In the world, there are about 1.3 billion people with vision impairment and 36 million people are blind according to the World Health Organization in 2018 [1]. Braille is an effective way for the visually impaired to learn knowledge, obtain information and communicate with other people.

Braille document consists of Braille characters and each Braille character has a rectangular block called Braille cell, which contains six Braille dots arranged in three rows and two columns with 64 different combinations [2]. Many Braille books are double-sided in order to save pages, which may contain recto dots and verso dots in one Braille image. Braille image recognition aims to automatically detect Braille dots in the whole Braille image and recognize Braille cells to Braille characters.

Many existing Braille image recognition methods are based on image segmentation and used several designed rules to discriminate the Braille dots [3, 4]. Some work used statistical learning methods to recognize Braille images [5, 6]. However, above Braille dots detection and Braille cell recognition methods are tested on their small-scale Braille images datasets with different acquisition ways and most of them are based on single-sided Braille. The performance on more complex, various Braille books and the large-scale dataset is lack.

There is no available public datasets for Braille image recognition to push relevant research and evaluate existing methods. This paper focuses on constructing such a large-scale Braille image dataset with double-sided, which can also provide Braille dots and Braille cells annotation information. And we also implement and evaluate some Braille dots detection methods on this dataset.

There are many ways to acquire Braille images from original Braille documents, including the camera, scanner and some special devices. Schwarz et al. [7] designed a controlled lighting environment to acquire Braille images with a fixed camera and fixed light sources, which is inconvenient and difficult to realize. Zhang et al. [8] used the camera of mobile phone to capture Braille images, which will be disturbed by background illumination and bring much distortion to captured images. Antonacopoulos et al. [3] used the flat-bed scanner to obtain Braille images, which is a quite simple and convenient way and with less image distortion. So we also adopt the general flatbed scanner to capture our Braille images.

The main contributions of this paper are as follows:

(1) We construct the first public Double-Sided Braille image dataset DSBI with Braille recto dots, verso dots and Braille cells annotation. This dataset includes 114 double-sided Braille images from 6 Braille books and some ordinary printed documents for keeping diversity and complexity. It is available on the Github website: https://github.com/yeluo1994/DSBI.

(2) An auxiliary annotation strategy is proposed to quickly annotate Braille images, which adopts initial automatic detection of Braille dots and modifies annotation results by convenient human-computer interaction method. This labeling strategy can averagely increase label efficiency by six times for recto dots annotation in one Braille image.

(3) Braille dots detection is the core and basic step for Braille image recognition. We evaluate some Braille dots detection methods on our Braille image dataset DSBI and give the benchmark performance of recto dots detection. These methods include Braille dots detection based on image segmentation and Haar+Adaboost classifier.

Our DSBI dataset is discussed in Section 2. Section 3 presents the proposed auxiliary annotation method. Experimental results analysis and conclusions are drawn in section 4 and section 5.

---

* Corresponding author

## 2. BRAILLE IMAGE DATASET

The purpose of constructing this dataset is to provide a benchmark for the current Braille recognition methods on real complex double-sided image. We will describe our construction method in details.

### 2.1 Braille Image Acquisition Way

As mentioned in the introduction section, there are many ways to acquire Braille images and different acquisition ways have different influence on the performance of Braille image recognition. We select flatbed scanner to obtain the double-sided Braille images, which is convenient and can provide good quality of Braille images. The scanner used in this paper is HP LaserJet Pro MFP M226dn. To reduce storing memory and remaining enough clarity, we use 200dpi resolution to capture color Braille images and store them in JPEG format. And we scan two sides of each Braille document to get recto dots and verso dots Braille information.

### 2.2 Braille Image Quality

Figure1 shows a local region sample of double-sided Braille image captured by our general HP scanner, which has uniform illumination. And it has good captured quality for Braille images recognition methods.

But in double-sided Braille images, some recto dots and verso dots are mixed together, which are difficult to achieve high-precision results of Braille image recognition. And to enhance diversity and complexity, our Braille images are acquired from several Braille books including different background document color, different production ways and different usage degree. Figure 2 shows some samples with above situation and some defects in our dataset, such as oil stains in Figure 2(a), paper distortion in Figure 2(b), some cracks in Figure 2(c), and abrasion Braille dots in Figure 2(d). These difference and defects are actually very common in real applications, which can also evaluate different methods with more objectivity.

### 2.3 Description of DSBI

Table 1 gives the detailed description of our constructed Braille image dataset DSBI. There are total 114 color double-sided Braille images from six different Braille books and six ordinary printed Braille documents.

These Braille books include reference books, such as Massage, professional textbooks, such as middle school Chinese textbook, and novels, such as Shaver Yang Fengting. To effectively evaluate Braille dots detection methods, we also add some pages containing only Braille verso dots in our dataset. And some Braille images have defects, such as oil stains from Massage book as Figure2 (a) shows. And the image quality from each book is also shown in Table 1.

For different Braille production ways and image capture noise, the Braille images we captured by scanner maybe exist some skewed angles. So our DSBI provides de-skewing Braille images and skewed angles for each image. And to quantitatively evaluate Braille image recognition performance, including Braille dots detection and Braille cells recognition, we provide corresponding Braille recto dots, Braille verso dots and Braille cells annotation information for each double-sided Braille image in our dataset DSBI.

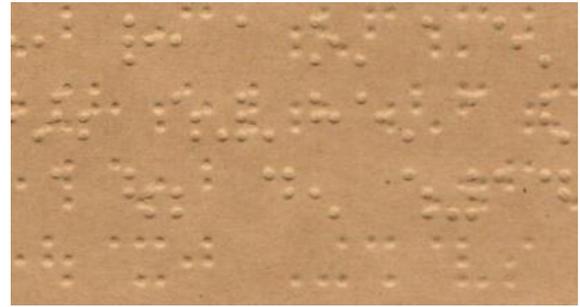

**Figure. 1 A region sample of double-sided Braille image.**

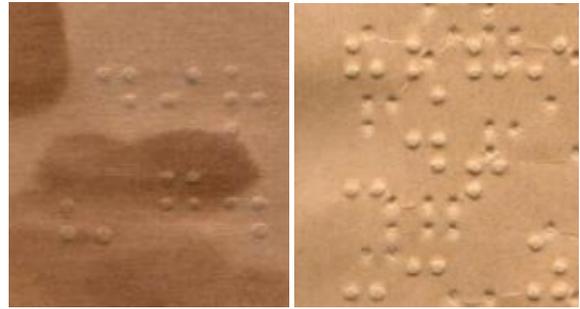

**(a) With oil stain**     **(b) With paper distortion**

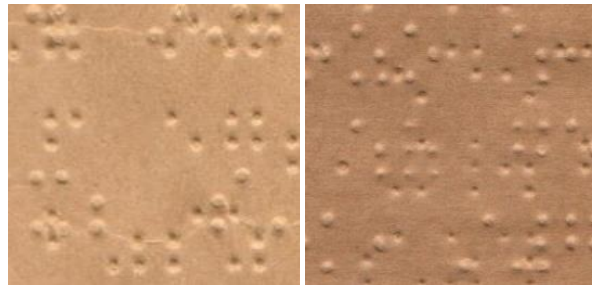

**(c) With some cracks**     **(d) With abrasion Braille dots**

**Figure 2. Complex samples of some local Braille images.**

**Table. 1: Detailed description of proposed dataset DSBI.**

| Book name | Total pages | Verso dots pages | Image quality |
|---|---|---|---|
| 1.Massage | 20 | 1 | Bad |
| 2.Fundamentals of Massage | 20 | 1 | Normal |
| 3.The Sec. Volume of Ninth Grade Chinese Book 1 | 20 | 1 | Normal |
| 4.The Second Volume of Ninth Grade Chinese Book 2 | 10 | 1 | Normal |
| 5.Math | 32 | 0 | Normal |
| 6.Shaver Yang Fengting | 6 | 0 | Normal |
| 7.Ordinary printed document | 6 | 0 | Good |
| Total | 114 | 4 | N/A |

## 3. AUXILIARY ANNOTATION METHOD

Accurate annotation of Braille dots and Braille cells for each Braille image in DSBI is important for developing recognition algorithms and evaluating performance. Our DSBI dataset contains double-sided Braille images and each image may contain about hundreds of recto dots or verso dots.

It is time-consuming to label each Braille dot and Braille cell information by a manual manner. We developed an interactive labeling tool, which will averagely cost over one hour to label only recto dots for one Braille image by the manual manner. For some Braille dots are too small to identify and some verso dots will seriously disturb the estimation of recto dots, which may bring many annotation errors. And manually labeling Braille cells information is a more complicated task. To reduce workload of labeling, this paper proposes a Braille image auxiliary annotation strategy as follows.

Firstly, we use the Haar+Adaboost and sliding windows strategy to automatically detect Braille recto dots in Braille images.

Secondly, the Braille dots may not be aligned in vertical and horizontal directions for production and capture noise. A Braille de-skewing process is used to correct Braille image. We get this skewed angle by analyzing the statistics information of row and column projections of detected Braille dots under different angles.

Thirdly, the distance between dots within a Braille cell and the distance between adjacent Braille cells are relatively fixed according to the acquisition resolution. We get the location of Braille cells based on above layout rules of Braille cells. We construct a Braille cell gird to generate preliminary Braille recto dots detection results.

Finally, based on above Braille recto dots and Braille cells location results, we propose a convenient interactive Braille annotation method using numeric keys and direction keys. Direction keys can quickly move Braille cell by Braille cell row and Braille cell column, and numeric keys can quickly modify the Braille dots information in each Braille cell to ensure the validity of annotation result. Figure 3 shows one sample of Braille image auxiliary annotation process.

For verso dots annotation, a recto dot on the front page is the verso dot on the back page for the double-sided Braille image. So we can easily obtain the verso dots annotation information by the recto dots annotation on the back page. Our Braille dataset DSBI provides the recto dots, verso dots and Braille cells location annotation information.

# 4. EVALUATION OF BRAILLE DOTS DETECTION METHODS

## 4.1 Evaluation Metrics

This paper used Precision, Recall and F1 value to evaluate the performance of Braille dots detection methods. The three indexes are as follows:

$$Precison = \frac{TP}{TP+FP} \quad (1)$$

$$Recall = \frac{TP}{TP+FN} \quad (2)$$

$$F1 = \frac{2 \times Precision \times Recall}{Precision+Recall} \quad (3)$$

Where TP represents the number of dots correctly identified, FN represents the number of dots misidentified as the background, and FP represents the number of background misidentified as dots.

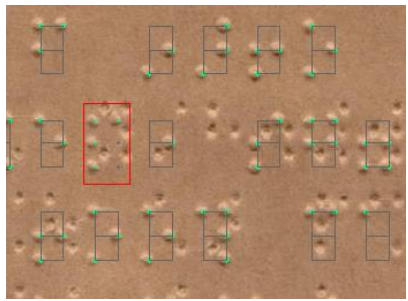

**Figure 3. A sample of Braille image auxiliary annotation.**

**Table. 2 Details of training set and test set of DSBI.**

| Book name | Training set pages | Test set pages |
|---|---|---|
| 1.Massage | 10 | 10 |
| 2.Fundamentals of Massage | 0 | 20 |
| 3.The Second Volume of Ninth Grade Chinese Book 1 | 0 | 20 |
| 4.The Second Volume of Ninth Grade Chinese Book 2 | 0 | 20 |
| 5.Math | 10 | 22 |
| 6.Shaver Yang Fengting | 3 | 3 |
| 7.Ordinary printed document | 3 | 3 |
| Total | 26 | 88 |

## 4.2 Training Set and Test Set

To debug the parameters of the segmentation method and obtain training samples for machine learning based methods, we divide our DSBI dataset into training set and test set. The total Braille images number of DSBI is 114, we select about 1/4 amount, 26 images, as training set and 3/4 amount, 88 images, as test set. Training images are from selected professional tool books, middle school textbooks, novels and ordinary printed documents. Test images are from each captured Braille book. The descriptive information of training set and test set is shown in Table 2.

## 4.3 Braille Dots Detection Methods

### 4.3.1 Based on image segmentation

Antonacopoulos et al. [3] used a local adaptive threshold to segment gray double-sided Braille image into three parts, including highlight, shadow and background, and obtained black and white region as Figure.4 shows. They divided the region whose width exceeded a threshold and then identified vertically an adjacent pair of white-black combination as recto dot, and black-white combination as verso dot.

We implement Braille recto dots detection method based on [3] and improve the original method. To reduce capture noise by Braille document edge for segmentation, we replace the unexpected pixel values with the Braille image background pixel value. Then gray normalization is adopted to optimize image quality and reduce the influence of different Braille images background. Gray histogram is used to find the adaptive global segmentation threshold for each Braille image. Finally, we identify vertically adjacent pairs of white-black regions to detect recto dots according to the size of white region and distance to the center of black region. This method is different from [3], which simply judges a recto dot according to whether a white region exists above the black region within the expected distance. We take the matched rectangle center as the position of recto dots as Figure.5 shows.

### 4.3.2 Based on Haar+Adaboost

The cascade classifier using Haar feature [9], frequently used in face detection and object detection, can get fast detection performance. This paper also used Haar+Adaboost method for Braille recto dots detection.

Haar feature can be calculated quickly by the integral image technology and cascade classifier. The cascade classifier can fast reject most of negative samples in the previous classifiers, which can greatly speed up the Braille dots detection speed. Meanwhile the cascade classifier has the advantage of generalization ability to ensure the high accuracy. We used Haar extraction and training function in OpenCV to train the cascaded classifier with selected positive samples and negative samples from training set based on annotation information. The sample size and sliding windows size are all 20×20 pixels, and the sliding step is 2 to get better performance.

## 4.4 Results and Analysis

In order to effectively compare the Braille dots detection methods in Section 4.3, we tested on the de-skewing Braille images in our test set from DSBI dataset. We calculate the Precision, Recall, F1 value of each method on the 88 double-sided Braille images as Table 3 shows.

It can be seen from the experimental results that the overall results of the two methods are relatively ideal. The Braille dots detection method based on image segmentation got 0.948 F1 value, which is the lowest among the methods. The segmentation based method will be subject to noise interference and need to set a lot of thresholds manually, which make it difficult to get better results in complex double-sided Braille images. The Braille dots detection method based on Haar+Adaboost got 0.97 F1 value, which is better than above image segmentation based method.

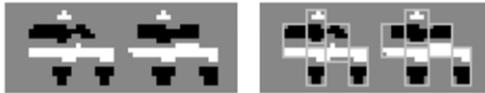

**(a) Segmented Braille image  (b) Corresponding dots**

**Figure 4. Segmented Braille Dots sample from [3].**

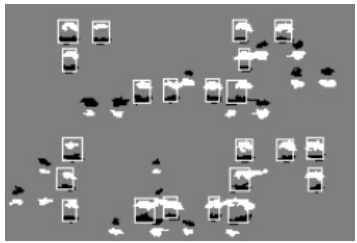

**Figure 5. Results of matched recto dots by segmentation.**

**Table.3 Comparison of Braille recto dots detection methods**

| Method | Precision | Recall | F1 |
|---|---|---|---|
| Based on Image segmentation | 91.72% | 98.11% | 0.948 |
| Based on Haar+Adaboost | 97.65% | 96.38% | 0.970 |

The Braille dots detection methods in this paper are carried out using c++ and OpenCV in the hardware environment of Intel Core I7 3.40 Ghz CPU and 16GB memory.

Above experimental results are based on Braille recto dots detection on the de-skewing double-sided Braille image. In our DSBI dataset, we also provide annotation information of verso dots and Braille cells, which can be used to test detection performance of verso dots and Braille cells. And besides de-skewing Braille images, we also provide the original double-sided Braille images and skewed angles information for each Braille image, which can be used for the de-skewing experiment.

## 5. CONCLUSION

This paper introduces our constructed Braille image dataset DSBI, which contains 114 double-sided Braille images from a variety of Braille books and documents. In addition, we also propose a Braille image auxiliary annotation method based on the automatic recognition of Braille images and modification of annotation results by convenient human-computer interaction method. Finally, we also implement the Braille dots methods based on image segmentation and Adaboost, and conduct experiments on our dataset DSBI to provide benchmark results. In future work, we will pay more attention to improving the performance of Braille dots detection and Braille cell location.

## 6. REFERENCES

[1] W. H. Organization, Visual impairment and blindness, 2018, [online] Available: http://www.who.int/en/news-room/fact-sheets/detail/blindness-and-visual-impairment.

[2] Isayed S, Tahboub R. A review of optical Braille recognition[C]//Web Applications and Networking, 2015 2nd World Symposium on. IEEE, 2015: 1-6.

[3] Antonacopoulos A, Bridson D. A robust Braille recognition system [C]//International Work- shop on Document Analysis Systems. Springer Berlin Heidelberg, 2004: 533-545.

[4] S. D. Al-Shamma and S. Fathi, "Arabic braille recognition and transcription into text and voice," in Biomedical Engineering Confer- ence (CIBEC), 2010 5th Cairo International. IEEE, 2010, pp. 227–231.

[5] Yin Jia. The key technology research on paper-mediated Braille automatic recognition system [Master degree thesis]. Changchun University of Science and Technology, Changchun, 2011 (in Chinese).

[6] Li Ting. A Deep Learning Method for Braille Recognition. Computer and Modernization, 2015, 36:37-40 (in Chinese).

[7] Schwarz T, Dolp R, Stiefelhagen R. Optical Braille Recognition[C]//International Conference on Computers Helping People with Special Needs. Springer, Cham, 2018: 122-130.

[8] Zhang S, Yoshino K. A braille recognition system by the mobile phone with embedded camera[C]//Innovative Computing, Information and Control, 2007. ICICIC'07. Second International Conference on. IEEE, 2007: 223-223.

[9] Viola. P. and Jones. M., "Rapid object detection using a boosted cascade of simple features", In IEEE Conference on Computer Vision and Pattern Recognition 2001.